\documentclass{article}

\usepackage[preprint]{neurips_2019} 
\setcitestyle{square,numbers,comma}

\usepackage[utf8]{inputenc} 
\usepackage[T1]{fontenc}    
\usepackage{hyperref}       
\usepackage{url}            
\usepackage{booktabs}       
\usepackage{amsfonts}       
\usepackage{nicefrac}       
\usepackage{microtype}      

\usepackage{tablefootnote}
\usepackage{graphicx}
\usepackage{amsmath}
\usepackage{amssymb}
\usepackage{subfigure}
\usepackage{amsfonts}
\usepackage{amssymb}
\usepackage{bbm}
\usepackage{amsthm}
\newtheorem{prop}{Proposition}

\usepackage[nomargin,inline,draft]{fixme}
\fxsetup{theme=color,mode=multiuser}
\FXRegisterAuthor{xg}{axg}{\color{red}xing}
\FXRegisterAuthor{pf}{apf}{\color{red}pascal}

\title{iPool - Information-based Pooling in Hierarchical Graph Neural Networks }

\usepackage{xcolor}
\author{
Xing Gao \thanks{Work performed at EPFL-LTS4,  visiting Prof. Pascal Frossard.}\\
Shanghai Jiao Tong University \\
Department of Electronic Engineering\\
\texttt{william-g@sjtu.edu.cn}
\And
Hongkai Xiong\\
Shanghai Jiao Tong University\\
Department of Electronic Engineering\\
\texttt{xionghongkai@sjtu.edu.cn}\\
\And
Pascal Frossard\\
\'Ecole Polytechnique F\'ed\'erale de Lausanne (EPFL)\\
Signal Processing Laboratory (LTS4)\\
\texttt{pascal.frossard@epfl.ch}
}

\begin{document}

\maketitle

\begin{abstract}
With the advent of data science, the analysis of network or graph data has become a very timely research problem. A variety of recent works have been proposed to generalize neural networks to graphs, either from a spectral graph theory or a spatial perspective. The majority of these works however focus on adapting the convolution operator to graph representation. At the same time, the pooling operator also plays an important role in distilling multiscale and hierarchical representations but it has been mostly overlooked so far. In this  paper, we propose a parameter-free pooling operator, called iPool, that permits to retain the most informative features in arbitrary graphs. With the argument that informative nodes dominantly characterize graph signals, we propose a criterion to evaluate the amount of information of each node given its neighbors, and theoretically demonstrate its relationship to neighborhood conditional entropy. This new criterion determines how nodes are selected and coarsened graphs are constructed in the pooling layer. The resulting hierarchical structure yields an effective isomorphism-invariant representation of networked data in arbitrary topologies. The proposed strategy is evaluated in terms of graph classification on a collection of public graph datasets, including bioinformatics and social networks, and achieves state-of-the-art performance on most of the datasets.
\end{abstract}

\section{Introduction}

Convolution neural networks (CNNs)  are efficient to extract hierarchical representations of signals residing on regular grids, such as audios and images.  With the convolution and  pooling operations, CNNs have achieved state-of-the-art performance in  a variety of applications.  With the increasing availability of various forms of network data, recent pioneer works \citep{bruna2013spectral,defferrard2016convolutional,khasanova2017graph,kipf2016semi,wang2018dynamic} have been generalizing convolution neural networks to irregular structures, including graph and point cloud data. Most of the current attempts for designing neural network representations of graph data however focus on the convolution operator. The pooling operator is mostly overlooked, yet it carries an important part of the ability of graph neural networks to distill effective hierarchical representations.

Hierarchical network data representations necessitate a careful design of all elements of the learning architecture. In tasks like graph classification for instance, a global representation is required in addition to local features in order to predict the label for an entire graph.  The pooling operator is an important component in the construction of such hierarchical architectures. In the case of image representation, the pooling operator downsamples data simply by flipping the predefined local receptive field and aggregating information in each receptive field, from left to right and top to bottom taking advantage of the inherent spatial order in lattice structures. However, in the general case of networks with  diverse irregular connections between vertices and no spatial order of the nodes, the pooling operation becomes more challenging. In particular,  it is not appropriate to directly generalize the pooling operator from images to graphs. Even for the simple pooling operation that downsamples image signals with stride 2, its counterpart on graphs that can be formulated as a max-cut problem, becomes a NP-hard problem \cite{bui1992finding}. In addition, it is still a challenge to construct a coarsened graph with the pooled features. In other words, one has to solve the following two problems to generalize the pooling operation to graphs: (1) How to choose and aggregate information to characterize the signals residing on graph? (2) How to coarsen the structure of graphs in the higher levels of representation? Graph theory may provide several tools to  generalize the pooling operator to networks with help of graph clustering or graph coarsening algorithms \citep{von2007tutorial,shuman2016multiscale,karypis1998fast}. However, although these methods perform well on fixed graph structures, they fail to easily generalize to networks with different topologies. Furthermore, their high computational complexity might become a limitation in graph convolution neural networks.

\begin{figure}
\centering
\includegraphics[width=5.5in]{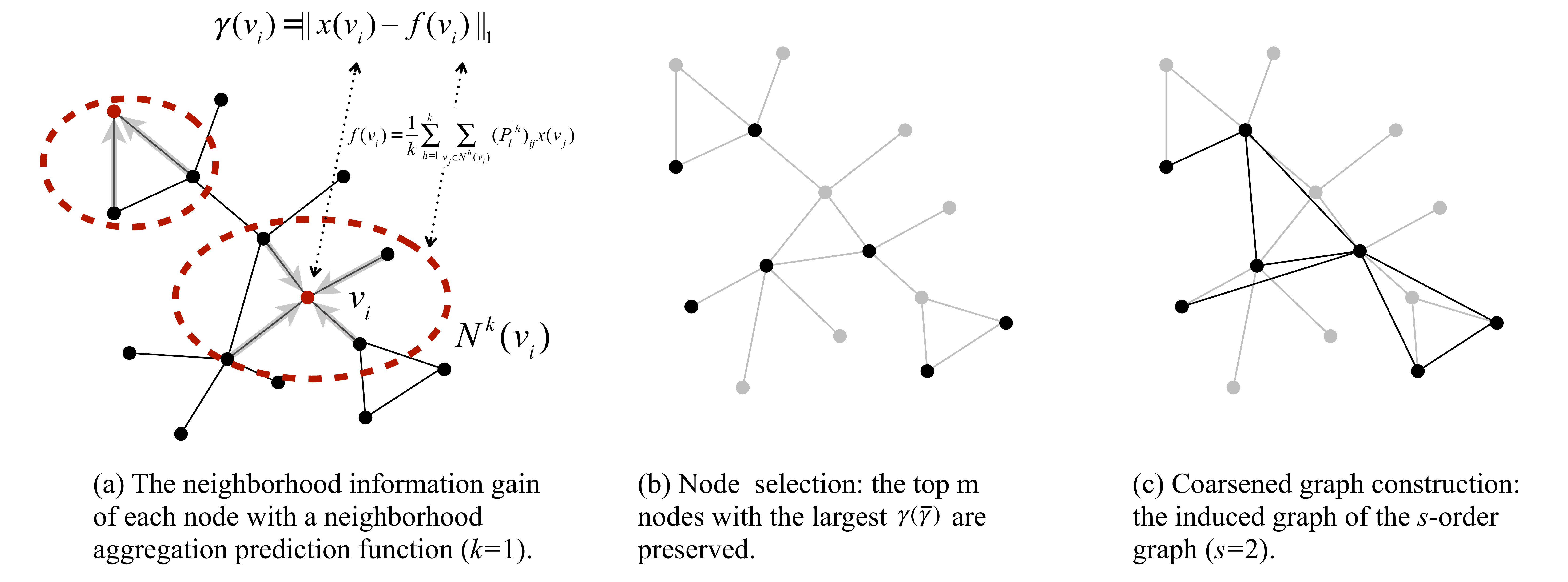}
\caption{Illustration of the iPool. The proposed neighborhood information gain criterion $\gamma(\cdot)$   and  prediction function $f(\cdot)$ are demonstrated in (a). The node selection strategy on the basis of the $\gamma$ ($\overline{\gamma}$) and coarsened graph construction are presented in (b) and (c), respectively. Specifically, $x(v_i)$ indicates the signal or feature on the node $v_i$, $N^k(v_i)$ represents the $k$-hop neighbors of the node $v_i$, and $\bar{{P_l}^h}$ implies an off-diagonal version of the $h$-th power of  transmission matrix.}
\label{fig:0}
\end{figure}

In this paper, we propose a pooling operator, called iPool, for data that comes with arbitrary graphs. This new operator is designed to generate a faithful representation of signals supported on the graph in terms of their information. Therefore, it first uses a criterion to evaluate the amount of information carried by each node given observations of its neighbors. Then, as informative nodes play an important role in characterizing graphs, we propose a strategy to construct coarsened graphs by aggregating information in accordance with the proposed information criterion, such as the intrinsic structure of the original graph is preserved. In particular, the node signals are predicted from the neighbour signals values to determine the most informative nodes, as shown in Fig.~\ref{fig:0}. A coarsened graph is constructed on these most informative nodes with a topology that maintains consistency with the original graph. The proposed iPool scheme offers several advantages: 1) it is a parameter-free building block, and it can be paired with diverse graph convolution operators in order to deal with arbitrary graphs; 2) it is invariant to graph isomorphism, which means that coarsened (isomorphic) graphs are identical through the pooling operation; 3) it exploits signals residing on graph in addition to the structure of graph; 4) it could provide two kinds of pooling schemes, namely global and local pooling which permit to trade off preservation of graph structure and information aggregation. Finally, we resort to graph classification tasks in order to evaluate the proposed method and we show that it outperforms state-of-the-art graph neural networks based methods on a collection of public benchmark datasets. The proposed pooling operator also has a promising perspective to be applied to other discriminative tasks and  generalized easily to other non-Euclidean data, such as point clouds.

\section{Related Work}
We give a brief review below on a collection of works on graph representation learning. We focus on traditional graph coarsening schemes as well as on recent attempts in graph pooling, which are relevant to the main problem addressed in this work. 

Graph neural networks (GNNs) are first proposed in \citep{gori2005new,scarselli2009computational,scarselli2009graph} and some recent works have attempted to extend  convolution neural networks to graphs, termed graph convolution networks (GCNs). One category of methods define convolutions for graphs on the basis of spectral filtering of graph signals, which leads to spectral GCNs \citep{bruna2013spectral,defferrard2016convolutional,khasanova2017graph,kipf2016semi}. Another line of works focus on defining the convolution operation in the spatial domain \citep{bruna2013spectral,hamilton2017inductive,wang2018dynamic,velikovi2018graph,zhang2018end,xu2018how} to address the basis-dependent problem where a spectral graph neural network trained on one graph structure fails to transfer properly to other graph structures. Most of them generalize the convolution operation by defining a ``message aggregation'' scheme, where information in the neighborhood is aggregated. These convolution layers in these architectures might be combined with different pooling operators to generate hierarchical representations of graph data.

When it comes to building multiscale representations of graph data, previous works rely on variants of graph coarsening algorithms, or differnet forms of graph pooling. For general graphs, unfortunately, this problem becomes a complex problem and a collection of approximate methods have been proposed in the literature, such as multiscale methods \citep{shuman2016multiscale,karypis1998fast}. Yet, computational complextiy stays high, so that these methods are mostly suitable for processing fixed structure graphs off-line, but have clear limitations in providing an on-line building block able to cope with graphs of various structures. An alternative to graph coarsening consists in implementing spectral clustering which split graphs into parts that can yield a hierarchical representation \citep{von2007tutorial}. Spectral clustering is however not ideal in building multiscale representations of graph data.

Pooling probably provides the most constructive alternative to develop multiscale representation in graph neural networks, and different methods have been proposed for generalizing poling ideas to graphs. For example \citeauthor{zhang2018end} \citep{zhang2018end} propose SortPooling that sorts the nodes based on the value of the last feature map in a descending manner, and preserves the first $k$ nodes of this list. On the other hand, DIFFPOOL \citep{ying2018hierarchical} follows clustering methods and assign nodes softly with generating the cluster assignment matrix using neural networks. SET2SET \citep{vinyals2015order} then implements an equivalent global pooling by aggregating information through LSTMs. However, these two recent methods have limitations: DIFFPOOL requires another graph neural network with the same capacity as the network for main task to learn projection matrices and the SET2SET depends on extra LSTM units which are time-consuming to train, in order to gather information. 

In contrary to previous methods, we propose a graph pooling scheme in accordance with the information of graph signals to enhance the capability of GNNs to construct hierarchical representations of graphs with arbitrary topologies. Similarly to the pooling operation for images, the proposed graph pooling operator is parameter-free  and ``plug and play'', and thereby leads to fast implementations in both training and testing phases.

\section{The iPool algorithm}

We present in this section the main elements of our new iPool algorithm. We first introduce the general framework of graph neural network architectures as well as the notation used in the paper. We later define the role of the pooling algorithm and introduce an information gain criterion that drives our iPool algorithm. We finally show how our pooling algorithm can be used to construct hierarchical representations with both global and local graph pooling strategies.

\subsection{Framework}

In this paper, we consider the learning of hierarchical representations for network data with help of a neural network architecture. Such an architecture is typically built on the concatenation of several layers, composed of graph convolution blocks and pooling operators. We focus here on the pooling operator, which is a key element for effective learning of hierarchical representations.

We define the notation used in this paper, where we employ capital letters and bold lowercase letters to indicate matrices and vectors, respectively. The network data at the input of the learning architecture, is represented by an undirected  graph $\mathcal{G}=(\mathcal{V},\mathcal{E})$, with $\mathcal{V}$ and $\mathcal{E}$ respectively the set of vertices, and the set of edges. The adjacency matrix $A$ represents the topology of the network, and presents a non-zero value at position $(i,j)$ (i.e., $A_{ij} \neq 0$) only if there is an edge in $\mathcal{E}$ that connects vertices $i$ and $j$. We consider both unweighted and weighted graphs, which leads to $A$ consisting of respectively unitary values or actual edge weights. We further define $D$ and  $P$ as the degree matrix and respectively  the transmission matrix  of $\mathcal{G}$. The degree matrix $D$ is a diagonal matrix with $D_{ii}=\sum_{j}A_{ij}$ and the transmission matrix $P$ defined as $P=D^{-1}A$ denotes the transmission probability of each pair of nodes. Furthermore, each vertex of the graph might further be attributed a signal value or feature, which is denoted by $\mathbf{x}_{i} = x(v_{i}) \in \mathbb{R}^d$, and $X=[\mathbf{x}_{1},\mathbf{x}_{2}, \cdots, \mathbf{x}_{n}]^{T}$. Finally, as we construct a hierarchical representation of graph data, the subscript $l$ is used to denote features or parameters belonging to the $l$-th layer of the neural networks. For example,  for the graph $\mathcal{G}_{l}=(\mathcal{V}_{l},\mathcal{E}_{l})$ in the $l$-th neural network representation layer, the $i$-th vertex of the graph is written as $v_{l,i} \in \mathcal{V}_{l}$, and $\mathbf{x}_{l,i} = x(v_{l,i})$ represents the signal or feature residing on the node $v_{l,i}$.

Most of current graph convolution networks utilize a stack of graph convolution layers to learn a representation or embedding of the graph data. The graph convolution layers are usually designed to follow a neighborhood message aggregation scheme, which can be written as:
\begin{equation}\label{e:0}
	X_{l+1}=\sigma(g(S_{l},X_{l},W_{l})),
\end{equation}
where $S_{l} \in \mathbb{R}^{n_{l}\times n_{l}}$ can be any graph shift operator that has nonzero values only in the positions corresponding to edges in the graph and in its diagonal, including but not limited to $A_{l}$ and $P_{l}$. The functions $g(\cdot)$ and $\sigma(\cdot)$ respectively denote a data aggregation function and a non-linear activation function, and the parameters $W_{l}$ are learned during the training of the neural network. The aggregation function $g(\cdot)$ varies in different graph convolution network architectures and is usually defined as  (weighted) summation \cite{ying2018hierarchical}, mean \cite{hamilton2017inductive, zhang2018end}, or even a Multilayer Perceptron (MLP) \citep{xu2018how}.  

A key element in these architectures is the graph pooling or coarsening operator, which aims at selecting a subset of data and obtain a version with a reduced dimension that still represents well the original graph data. These operators equip the neural networks with the ability to construct a sort of multiscale representations of network data. However, due to the diverse structures of graphs and the absence of a regular grid-like topology in general, it is not possible to pre-define the sampling structure or the local receptive fields on graphs, as it can be done in image representation learning \cite{bui1992finding}. The graph pooling operator has thus to be defined adaptively, yet in a generic way so that it can accommodate different network structures. 

Generally, for an undirected graph $\mathcal{G}_{l}=(\mathcal{V}_{l},\mathcal{E}_{l})$, the graph pooling process could be formulated as:
\begin{equation}\label{e:2.1}
	A_{l+1}=C_l \ A_l \ {C_l}^{T},
\end{equation} 
where $A_{l+1} \in \mathbb{R}^{n_{l+1} \times n_{l+1}}$ and $A_l \in \mathbb{R}^{n_{l} \times n_{l}}$ are the adjacency matrices of the coarse and fine graphs and $C_l \in \mathbb{R}^{n_{l+1} \times n_{l}}$ is the coarsening matrix. Correspondingly, the signal $X_{l}$ residing on the graph is downsampled by the coarsening matrix:
\begin{equation}\label{e:3}
	X_{l+1}=C_l X_l.
\end{equation} 
Notably, Eq.(\ref{e:2.1}) and Eq.(\ref{e:3}) are general formations that are suitable for both traditional graph coarsening as well as clustering methods \cite{kushnir2009efficient, chevalier2009comparison,kushnir2006fast} and state-of-the-art graph pooling methods \cite{ying2018hierarchical}.

The design of effective hierarchical representations of graph data largely relies on the proper choice of the graph pooling operator, hence on a proper choice of the operator $C$. We present below a novel pooling strategy using a neighborhood information gain criterion. 

\subsection{The neighborhood information gain criterion}

In order to design a proper pooling operator, one first needs to define a criterion that governs the selection of the most important nodes in the graph. The objective is to coarsen the graph representation while keeping a faithful representation of the original one, with the coarsening matrix or equivalently the pooling operator generated on the basis of the structure of graph and the signals that it supports. In general, if a signal residing on one particular node of the graph could be well predicted from signals supported by other nodes, this node can probably be removed in the coarsened graph, with negligible information loss. If we further consider the typical localization and smoothness properties of most signals, it is reasonable to limit the node signal prediction process within the node neighborhood. Therefore, we can relate the amount of information carried by a graph node, to the difficulty of predicting the signal value from nodes in its neighborhood. We therefore introduce below a measure, namely the neighborhood information gain criterion, in order to quantitatively evaluate the uncertainty or information of node signals given observations of the neighbors. We later use this measure to design a new pooling operator that eventually preserves the most important nodes in the graph. 

The neighborhood information gain criterion is defined as the ``Manhattan'' distance between the observed signal $\mathbf{x}_{l,i} = x(v_{l,i})$ and the one predicted from observations at the neighbor nodes \footnote{For graph without signals, graph structure is taken as signals  by using features after convolution layers with constant features as input (i.e., $\mathbf{1}$ for all nodes) implicitly.}. We choose the ``Manhattan'' distance as it represents a common similarity measure that is especially convenient for high dimensional vectors that might be present in some graph datasets. Specifically, with a prediction function $f(\cdot)$ using information from the neighborhood, the neighborhood information gain of each node could be formulated as:
 \begin{equation}
\label{e:1}
\gamma(v_{l,i})	=\parallel x(v_{l,i}) - f( v_{l,i} )\parallel_{1}.
\end{equation}
With this definition, the neighborhood information gain criterion  gives the same importance to differences along  each dimension; using an $\ell_1$ norm prevents specific dimensions with large deviations from dominating the other ones as it typically happens with $\ell_p$ norms with $p>1$. We now have different options for choosing the prediction function $f(\cdot)$. Among them, neighborhood aggregating functions are promising, in particular when considering the typical localization and smoothness properties of graph signals. We therefore choose to predict the node signal as the weighted average of signals supported on nodes within its $k$-hop neighborhood.  Given that the $h$-th power of transmission matrix $ {P_{l}}^{h}$ is a effective measure of the level of connection or dependence between any pair of nodes reachable with $h$ hops, we adopt the elements of $ {P_{l}}^{h} (h=1, 2, \dots, k)$ as the weights in our prediction function $f$ in order to give more confidence to nodes that have stronger connections. We however modify ${P_{l}}^{h}$ into an off-diagonal transition matrix $\bar{{P_{l}}^{h}}$ corresponding to graph without $h$-hop circles. Finally, the prediction function can be formulated as
\begin{equation}\label{e:2}
	f(v_{l,i})=\frac{1}{k}\sum_{h=1}^{k}\sum_{v_{l,j} \in N^{h}(v_{l,i})}(\bar{{P_{l}}^{h}})_{ij}x(v_{l,j}),
\end{equation}
with
\begin{equation}
	\bar{{P_{l}}^{h}}=(\bar{{D_{l}}^{h}})^{-1}\bar{{A_{l}}^{h}}, \quad \bar{{A_{l}}^{h}}={A_{l}}^{h}-{\rm diag}({A_{l}}^{h}),
\end{equation}
where $N^{h}$ is the $h$-hop neighborhood, $\bar{{A_{l}}^{h}}$ is the adjacency matrix where diagonal values corresponding to the $h$-hop circles have been removed, and $\bar{{D_{l}}^{h}}$ is the corresponding degree matrix. In this way, Eq. ~(\ref{e:1}) could be further formulated as:
\begin{gather}
    \gamma(v_{l,i})	=\parallel x(v_{l,i})-\frac{1}{k}\sum_{h=1}^{k}\sum_{v_{l,j} \in N^{h}(v_{l,i})}(\bar{{P_{l}}^{h}})_{ij} \times x(v_{l,j}) \parallel_{1}, \\
	\Gamma(\mathcal{G}_{l})=\mid (I-\frac{1}{k}\sum_{h=1}^{k}{\bar{{P_{l}}^{h}}})X_{l}\mid,
\end{gather}
where we use $\mid \cdot \mid$ to indicate the $l_{1}$ norm of each row of a matrix. The value of $\gamma$ will be high when a node signal is very different from the ones in its neighborhood, which means that the node has high information and should be preserved in the pooling operator. Note that the definition of the information criterion is local, which is very important towards low complexity and possible distributed implementation.

\subsection{Graph Coarsening}

With the neighborhood information gain criterion defined above, we can now identify nodes that should be preserved by the pooling operator. The convolution layers based on message aggregation (see Eq.~(\ref{e:0})) performs a series of equivalent low-pass filtering of the graph information. The pooling then identifies the nodes that have the higher information gain and permits to adaptively downsample the graph while preserving the local characteristics of graph signals. We describe below a local and global version of the pooling algorithm, and then show how we construct coarsened graphs after pooling in order to build a multiscale representation of network data. 

Similarly to pooling strategies developed for image representations, a global pooling scheme can aggregate information of the whole graph globally at the possibly expense of losing the graph structure information, while local pooling strategy can preserve the general structure of graph with however a limited receptive field to collect information. With different applications having diverse requirements, it is worthwhile to explore both of global pooling and local pooling for graphs. Luckily, with the neighborhood information criterion defined above, we can derive both global pooling as well as local pooling strategies with only minor adjustments.

\textbf{Global iPool strategy.} On the basis of the neighborhood information gain, nodes are assigned different priorities to construct coarse graph globally. In order to approximate the information of the graph, the pooling should preserve the nodes that can not be well represented by their neighbors. In other words, the nodes with relative high neighborhood information gain have to be preserved in the construction of a coarsened graph. Specifically, the graph nodes are re-ordered based on the value of their neighborhood information gain. The global pooling strategy then adopts a common step as \citep{zhang2018end,DBLP:conf/icml/GaoJ19} that simply selects the top $n_{l+1}=\rho \times |\mathcal{V}_{l}|$ nodes that have the highest information gain, as:
\begin{equation}
	{\rm \mathbf{idx}}={\rm rank}(\Gamma(\mathcal{G}_{l}), n_{l+1}),
\end{equation}
where $\rho$ is the pooling ratio and $\rm rank$ represents the global ranking operator. 

\textbf{Local iPool strategy.} Pooling can also be implemented locally, and nodes can be selected within each receptive field, similarly to what is done for images. Receptive fields may however overlap, and we propose to normalize the information gain in each neighborhood before local pooling. This reduces the probability that selected nodes mainly come from dominant neighborhoods and permits to have a better distribution of the pooled nodes over the original graph. Mathematically, the neighborhood information gain of each node is normalized by the average neighborhood information gain of its neighborhood:
\begin{equation}
\overline{\gamma}(v_{l,i})=\frac{\gamma(v_{l,i})}{\sum \limits_{v_{l,j} \in  N(v_{l,i})}(P_{l})_{ij}\gamma(v_{l,j})}.
\end{equation}
Then, the nodes are ordered and selected globally in terms of the normalized neighborhood information gain:
\begin{equation}
	 {\rm \mathbf{idx}}={\rm rank}(\overline{\Gamma}(\mathcal{G}_{l}), n_{l+1}).
\end{equation}

\textbf{Coarsened graph construction.} The above iPool versions select the nodes to be preserved and we can now construct a coarsened graph with the selected nodes. First, the coarsening matrix of Eq.~(\ref{e:2.1}) and Eq.~(\ref{e:3}) can be obtained directly from the indices of the nodes selected by iPool. Concretely, with the indices of the pooled nodes, the coarsening matrix $C_{l} \in \mathbb{R}^{n_{l+1} \times |\mathcal{V}_{l}|}$ is constructed according to the following formulation,  for $\forall i ~\in ~\{1,2,\dots,n_{l+1}\}$:
\begin{equation}\label{e:14}
(C_l)_{ij}=\left\{
\begin{array}{ll}
1  \quad j={\rm \mathbf{idx}}(i) \\
0  \quad  {\rm  others. }
\end{array}\right.	
\end{equation}
The coarsening matrix is sparse; the cardinality of each row is $1$, and the cardinality of each column is no more than $1$. 

The coarsened graph is constructed as follow. To facilitate propagation of signals on the graph,  we first alter the connection between nodes in the original  graph, similar to \cite{DBLP:conf/icml/GaoJ19}. Specifically,  there is an edge connecting two nodes if and only if there is a path consisting of $s$ edges between these two nodes in the original graph. The weights of the edges are further set  either   as the corresponding elements of the $s$-th power of the  adjacency matrix for weighted graphs or as  $1$ for unweighted graphs. Namely,  we set
\begin{equation}
	(\tilde{A}_l)_{ij}=1 \iff ({A_{l}}^s)_{ij} > 0, \quad  i\neq j,
\end{equation}
for unweighted graphs.  For weighted graphs, we set
 \begin{equation}\label{e:edge}
	   {\tilde{A}}_{l} ={A_{l}}^{s}-{\rm diag}({A_{l}}^{s}),
\end{equation}
where  the value of $s$ is set to be either 1 or 2 depending on the properties of the datasets and applications. Notably, the hyper-parameter $s$ plays a similar role as the hyper-parameter stride in image pooling. The coarsening matrix in Eq.~(\ref{e:14}) is applied to the adjacency matrix of the expanded graph ${\tilde{A}}_{l}$ to obtain the coarsened graph. The graph and features at the next layer of the representation thus become:
\begin{gather}
	A_{l+1}=C_{l}{\tilde{A}}_{l}{C_{l}}^{T},\\
	X_{l+1}=C_{l}X_{l}.
\end{gather}

Note that, during the training phase of the neural network architecture, the iPool operator as any general end-to-end differentiable layer, takes as inputs the adjacency matrix and node features of graphs and produces the adjacency matrix and node features of coarsened graphs. Furthermore, the operator is generic enough to be integrated in diverse architectures.

\begin{figure}[tp]
\centering
\includegraphics[width=5.5in]{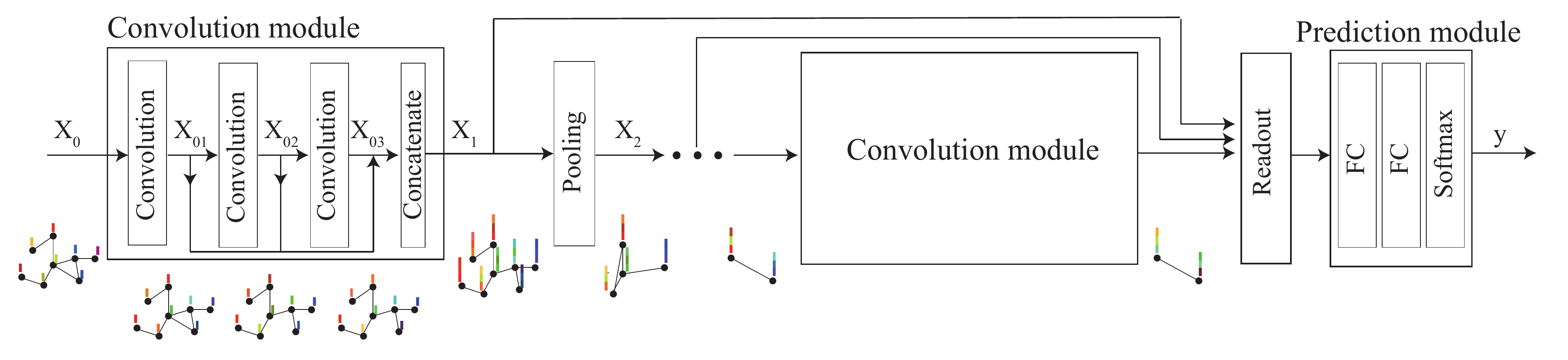}
\caption{  Hierarchical graph convolution network architecture. The network is composed of a stack of convolution modules, iPool layers, a readout module as well as a prediction module. }
\label{fig:1}
\end{figure}

\section{Properties of iPool}
\label{sec:ipool_properties}

In this section, we will discuss the relationship between the neighborhood information gain and neighborhood conditional entropy and derive some properties of iPool.

The neighborhood information gain is defined as the deviation between the observed signal and the predicted signal based on signals in the neighborhood. Empirically, this deviation reflects the uncertainty of one signal value given other values in its neighborhood. In information theory, the entropy is designed to quantify the uncertainty, and the conditional entropy is specifically  utilized to measure the amount of information of one variable given values of the other variable(s). Generalized to graph signals, we arrive at the neighborhood conditional entropy by considering variables as signals supported on nodes in neighborhood: 
\begin{equation}
	H(\mathbf{x}_{l,i}| \{\mathbf{x}_{l,j}\}_{  N(v_{l,i})})=H(\mathbf{x}_{l,i})-I(\mathbf{x}_{l,i};\{\mathbf{x}_{l,j}\}_{  N(v_{l,i})}),
\end{equation}
where $H(\cdot)$ is the entropy and $I(\cdot)$ represents the mutual information.
Actually, under a certain assumption on the conditional distribution, the proposed neighborhood information gain has a close relationship with the neighborhood conditional entropy.
\begin {prop}\label{p:0.1} 
	Let assume that the components of  neighborhood conditional distribution of each node are independent and that each component satisfies a Laplace distribution  $p(x_{l,i,z}| \{\mathbf{x}_{l,j}\}_{  N(v_{l,i})})$=  Laplace($\mu_{l,i,z}, b_{l,i}$) with $\mathbf{\mu}_{l,i}=[\mu_{l,i,1},\mu_{l,i,2}, \dots, \mu_{l,i,d}]=	f(v_{l,i})$, the neighborhood information gain $\gamma(v_{l,i})$ of each node is an approximate empirical estimation of its neighborhood conditional entropy. 
\end {prop}

With Prop.~\ref{p:0.1}, the proposed iPool algorithm is a constructive way to coarsen graphs in accordance with \textit{the maximum neighborhood conditional entropy strategy}. Specifically, the global iPool preserves nodes with the maximum neighborhood conditional entropy, while the local iPool selects nodes with relatively large neighborhood conditional entropy. Furthermore, the global iPool strategy implies that nodes share the same or similar variations of neighborhood conditional distribution, while the local iPool strategy relaxes this restriction to nodes in the same neighborhood having similar variations in terms of neighborhood conditional distribution. The proof  is presented in Appendix.

We now further show that iPool is invariant to isomorphic graphs so that the graph neural networks consisting of iPool combined with other components that are also invariant to graph isomorphism, will produce invariant representations for isomorphic graphs. This property is important for a wide collection of discriminative tasks, such as graph classification.  We prove this proposition in Appendix.

\begin {prop}\label{p:3}
For any isomorphic graphs  $\mathcal{G}_{l}=(\mathcal{V}_{l},\mathcal{E}_{l},X_{l})$ and $\mathcal{G}_{l}^{'}=(\mathcal{V}_{l}^{'},\mathcal{E}_{l}^{'},X_{l}^{'})$, the iPool will produce the same coarsened graphs. 
\end {prop} 

In summary, the information gain criterion used in iPool is actually a constructive approximation of the maximum neighborhood conditional entropy, which further validates the choice of this measure to preserve the graph information.  Furthermore,  the iPool algorithm is invariant under graph isomorphism, which is very important in numerous tasks.

\section{Experiments}
In this section, we evaluate the effectiveness of the proposed pooling scheme in graph classification tasks, and compare with a variety of kernel based methods and neural network based methods.

\subsection{Experimental settings} 
We evaluate the proposed pooling algorithm in the context of  deep graph convolution networks. The hierarchical graph convolution networks used in the experiments following the general framework of GraphSAGE \cite{hamilton2017inductive} consist of a stack of convolution modules, iPool layers, a readout module as well as a prediction module, as presented in Fig.~\ref{fig:1}. 

\begin{table}[tp]
\centering
 \caption{Graph classification accuracies with 10-fold cross-validation with the large networks. Results of baseline methods with `*' and `$\dagger$' are respectively cited from \cite{ying2018hierarchical} and the respective publications. }\label{t:1}
\begin{tabular}{l|ccccc}
	\toprule
	Method & ENZY & D\&D & RED-M-12K & COLL & PROT\\
    \midrule
	GRAPHLET$^{*}$ & 41.03 & 74.85 & 21.73 & 64.66 & 72.91 \\
	SP$^{*}$& 42.32&78.86&36.93&59.10&76.43\\
	WL$^{*}$ &53.43&74.02 & 39.03&78.61 &73.76\\
	WL-OA$^{*}$&\textbf{60.13} &79.04 &44.38&\textbf{80.74}&75.26\\
	\midrule
	PSCN$^{\dagger}$& -&76.27$\pm$2.64&41.32$\pm$0.42&72.60$\pm$2.15&75.00$\pm$2.51\\
	GRAPHSAGE&54.00$\pm$4.36&78.08$\pm$3.52&41.19$\pm$7.19&76.28$\pm$1.67&76.55$\pm$4.20\\
	ECC$^{\dagger}$&53.50&74.10&41.73&67.79&72.65\\
	SET2SET&55.50$\pm$6.99&78.51$\pm$3.59&> 7days&75.46$\pm$1.40&76.73$\pm$3.97\\
	SORTPOOL$^{\dagger}$&57.12&79.37&41.82&73.76&75.54\\
        DIFFPOOL&58.67$\pm$5.37 &79.19$\pm$4.17&47.37$\pm$1.02&76.38$\pm$1.93&77.00$\pm$2.93\\
\midrule  
		Proposed global&56.00$\pm$7.72&78.76$\pm$3.45&47.02$\pm$1.01&76.86$\pm$1.67&76.46$\pm$3.22 \\
	Local ($k=1$)  &\textbf{59.00$\pm$5.73}&\textbf{79.45$\pm$2.78}&47.24$\pm$1.10&\textbf{77.28$\pm$2.17}&\textbf{77.36$\pm$3.27} \\
	Local ($k=2$) &57.50$\pm$6.20&78.93$\pm$4.07&\textbf{47.64$\pm$1.56}\tablefootnote{Better performance will be achieved with other configurations, and detailed information is presented in Ablation studies.}& 77.20$\pm$1.76&76.72$\pm$3.06\\
	\bottomrule
	
\end{tabular}
 \end{table}

\begin{table}
\centering
 \caption{  Dataset statistics and properties. }\label{t:a2}
 \resizebox{\textwidth}{12mm}{
\begin{tabular}{l|cccccccccc}
	\toprule
	Method & ENZY & D\&D & RED-M-12K & COLL & PROT & NCI1 & NCI109 & MUTAG  &IMDB-B&IMDB-M\\
	\midrule
		Avg $|\mathcal{V}|$ &32.63 &284.32&391.41&74.49&39.06&29.87&29.68&17.93&19.77&13.00\\
		Avg $|\mathcal{E}|$ &62.14 &715.66&456.89&2457.78&72.82&32.30&32.13&19.79&96.53&65.94\\
		Node labels &V &V&X&X&V&V&V&V&X&X\\
		\#Classes&6 &2&11&3&2&2&2&2&2&3\\
		\#Graphs & 600&1178&11929&5000&1113&4110&4127&188&1000&1500\\
	\bottomrule
\end{tabular}}
\end{table}

The convolution module consists of three graph convolution layers. For the graph convolution layer, we follow the design of common architectures and choose the general message propagation and aggregation formation. Specifically, we adopt the following convolution module:
\begin{equation}\label{e:a0}
	X_{l+1}=\text{ReLU}(\phi(A_{l}X_{l}W_{l})),
\end{equation}
where ReLU denotes the rectified linear unit activation function and $\phi(\cdot)$ indicates an $l_{2}$ normalization function to stablise and accelerate the training process.  In more details, this convolution operation consists of three steps: (1) Neighborhood information aggregation: through $A_{l}X_{l}$,  each node propagates information in neighborhood and make a weighted summation of  information from neighbors utilizing the weights of shift matrix $(A_l)_{ij}$.  It is analogous to weighted average filtering, a common low-pass filtering,  and thereby enables the graph convolution networks to be robust to noisy signals to some degree. (2) Affine transformation: if we unfold the formation $\tilde{X_{l}}=[\mathbf{\tilde{x}}_{l,1},\mathbf{\tilde{x}}_{l,2}, \cdots, \mathbf{\tilde{x}}_{l,n}]^{T}=A_{l}X_{l}=A_{l}[\mathbf{x}_{l,1},\mathbf{x}_{l,2}, \cdots, \mathbf{x}_{l,n}]^{T}$,  we will see that  $\tilde{X}_{l}W_{l}$ imposes an affine transformation on each node after aggregation, because of  $\tilde{X_{l}}W_{l}=({W_{l}}^{T}\tilde{{X}_{l}}^{T})^{T}=({W_{l}}^{T}[\mathbf{\tilde{x}}_{l,1},\mathbf{\tilde{x}}_{l,2}, \cdots, \mathbf{\tilde{x}}_{l,n}])^{T}$. (3) Non-linear mapping: the ReLU non-linear function is chosen to further perform point-wise non-linear transformation.  Feature maps of convolution layers within a convolution module are further concatenated to form the outputs of the convolution module. 
The pooling module then follows the convolution module to coarsen graphs in accordance with the iPool operator, as introduced in Section 3. In addition to convolution and pooling modules,  a readout module is utilized to attain graph embeddings of different coarsened versions and these graph embeddings are concatenated to produce the final graph representation:
\begin{equation}
    h_{\mathcal{G}}={\rm Concat}(\eta( X_{l})| l=1,3,\dots, K),
\end{equation}
where $\eta(\cdot)$ indicates an element-wise operator to aggregate information of all nodes along each dimension of features. Specifically, an element-wise sum operator is used on biological datasets and an element-wise mean operator is adopted on social network datasets. A prediction module is finally added to the architecture for graph classification. It consists of two fully connected layers and a soft-max layer to make the prediction of the graph category based on the graph representation $h_{\mathcal{G}}$.

\begin{table}[tp]
\centering
\caption{Graph classification accuracies with 10-fold cross-validation with the small networks.  Results of baseline methods with `$\dagger$' are  cited from original publications.}\label{t:2}
 \begin{tabular}{l|cccccc}
	\toprule
	Method & NCI1 & NCI109 & MUTAG  &IMDB-B&IMDB-M\\
    \midrule  
    PSCN$^{\dagger}$ &76.34$\pm$1.68 &- &88.95$\pm$4.37& 71.00$\pm$2.29&45.23$\pm$2.84\\
	GRAPHSAGE  &78.76$\pm$1.73&77.92$\pm$2.47  &86.78$\pm$8.88 &72.10$\pm$2.66&49.20$\pm$3.94\\
	ECC$^{\dagger}$ & 76.82 &75.03 &76.11  &-&-\\
	SORTPOLL$^{\dagger}$ &74.44 &- &85.83 &70.03&47.83&\\
	k-GNNs$^{\dagger}$&76.2&-&86.1&\textbf{74.2}&49.5\\
    SET2SET&80.80$\pm$1.72&79.09$\pm$1.96&86.78$\pm$7.33&71.00$\pm$7.54&49.73$\pm$4.19\\
    DIFFPOOL &80.46$\pm$1.22 &79.04$\pm$1.25&88.87$\pm$6.75 &73.00$\pm$3.22&49.60$\pm$3.87\\
\midrule 
Proposed global &80.46$\pm$1.66&78.80$\pm$2.62&89.42$\pm$5.68&72.90$\pm$3.08&50.73$\pm$3.68\\
Local ($k=1$)    &81.41$\pm$1.53&80.01$\pm$2.32&87.84$\pm$6.12&73.10$\pm$2.98&50.53$\pm$2.71\\
Local ($k=2$)    &\textbf{81.58$\pm$1.46}&\textbf{80.03$\pm$2.05}&\textbf{90.42$\pm$4.68}&\textbf{73.30$\pm$2.72}&\textbf{51.27$\pm$3.44}\\
	\bottomrule
\end{tabular}
 \end{table}

 We implement the proposed model in Pytorch \citep{paszke2017automatic}.  We conduct experiments to classify graphs on ten public benchmark graph datasets, including  biological datasets (MUTAG, ENZYMES, DD, PROTEINS, NCI1, NCI109) and social network datasets (IMDB-BINARY, IMDB-MULTI, COLLAB,  REDDIT-MULTI-12K) \footnote{Datasets could be downloaded from https://ls11-www.cs.tu-dortmund.de/staff/morris/graphkerneldatasets}. Statistics and properties of the datasets are presented in Table~\ref{t:a2}. Node categorical features are adopted in biological datasets, while constant features are utilized in social datasets.  Following prior methods \citep{ying2018hierarchical,xu2018how}, we perform  10-fold cross-validation on all of the datasets and report the best average accuracy. In the experiments, we employ two kinds of networks,  a small one with the number of  hidden neurons  as 30   and a large one with 64 hidden neurons at  each convolution layer. Also, one iPool layer is utilized for most  datasets except that two iPool layers are employed for D\&D and REDDIT-M-12K because of the large number of nodes per graph.  The pooling ratio ($\rho$)  is set as $0.1$ for middle datasets, including ENZYMES,  PROTEINS, and COLLAB, and $0.25$ for the others. These networks are optimized by mini-batch gradient descent algorithm (batch size =20) with the Adam  optimizer. The following hyper-parameters are tuned for each dataset: (1) learning rate $\in \{0.01, 0.001, 0.0001\}$;  (2) the drop-out ratio $\in \{0, 0.5\}$ and weight decay $\in \{0, 3e-5, 1e-4\}$.  Detailed information about hyper-parameters is presented in Appendix (Table~\ref{t:a1}).  Finally, we compare the proposed method with  a collection of state-of-the-art kernel-based solutions as well as GNN-based graph representation learning methods.  For graph kernel methods, we compare with  the Weisfeiler-Lehman subtree kernel (WL) \cite{shervashidze2011weisfeiler}, the Weisfeiler-Lehman optimal assignment kernel (WL-OA) \cite{kriege2016valid}, graphlet kernel \cite{shervashidze2009efficient}, and shortest-path kernel (SP) \cite{borgwardt2005shortest}. On the other hand,  a series of graph neural network variants and different graph pooling schemes designed for deep graph neural networks are taken into consideration, including  PSCN \cite{niepert2016learning}, GraphSAGE \cite{hamilton2017inductive},  ECC \cite{simonovsky2017dynamic},  k-GNNs \cite{morris2019weisfeiler}, global pooling scheme with LSTMs (SET2SET) \cite{vinyals2015order}, sort pooling (SORTPOOL) scheme  \cite{zhang2018end}, and DIFFPOOL generating coarsening matrix with an extra GNN \cite{ying2018hierarchical}. Results of the baseline models are cited from the original works except the results of GraphSAGE, SET2SET, and DIFFPOOL methods which are obtained using their public code with the same architecture used for iPool.

\begin{figure}[tbp]
\centering
\subfigure[Global pooling ($k=2$).]{
\begin{minipage}{0.48\linewidth}
\centering
\includegraphics[width=2.5in]{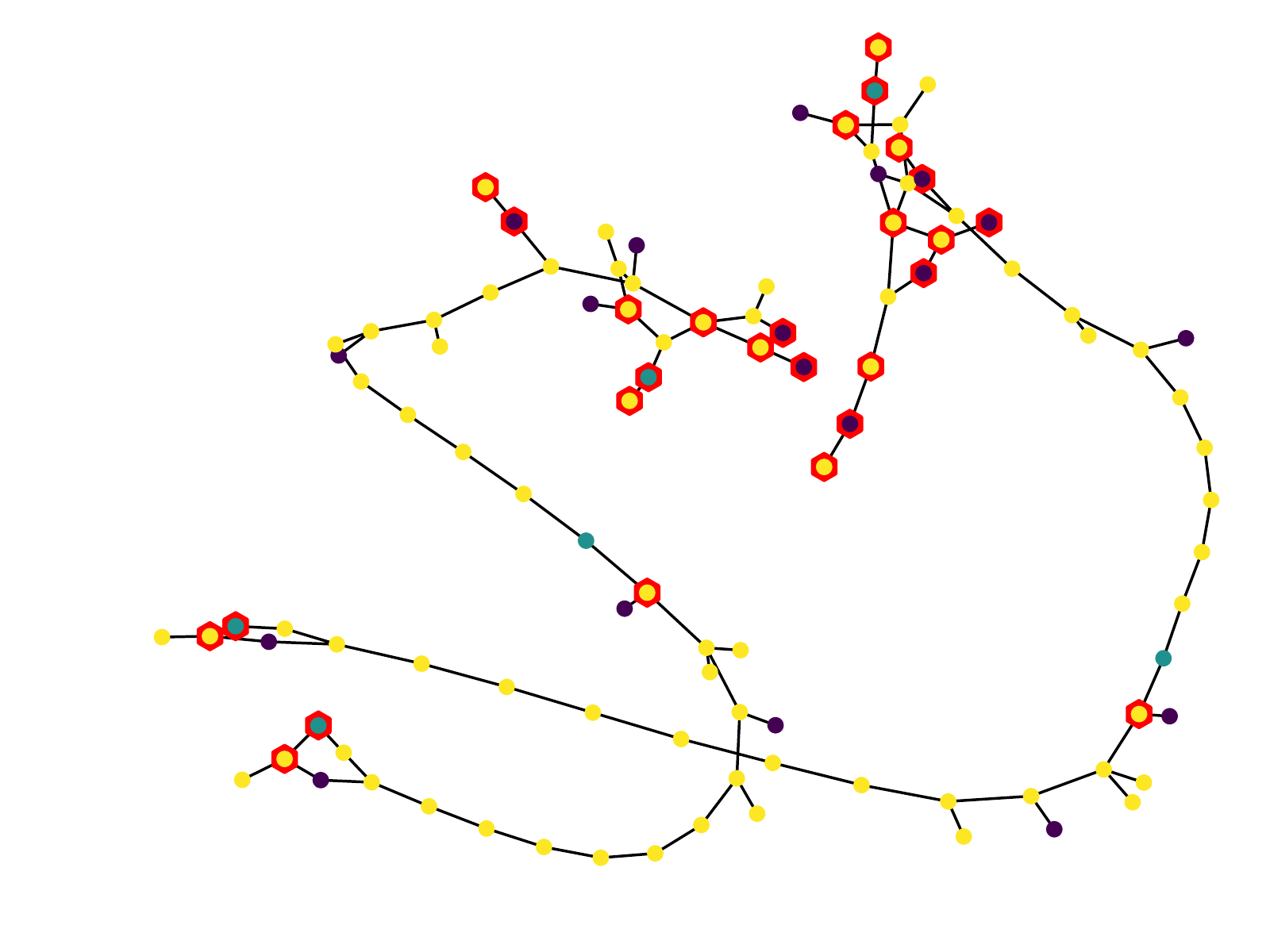}
\end{minipage}
}
\subfigure[Local pooling ($k=1$).]{
\begin{minipage}{0.48\linewidth}
\centering
\includegraphics[width=2.5in]{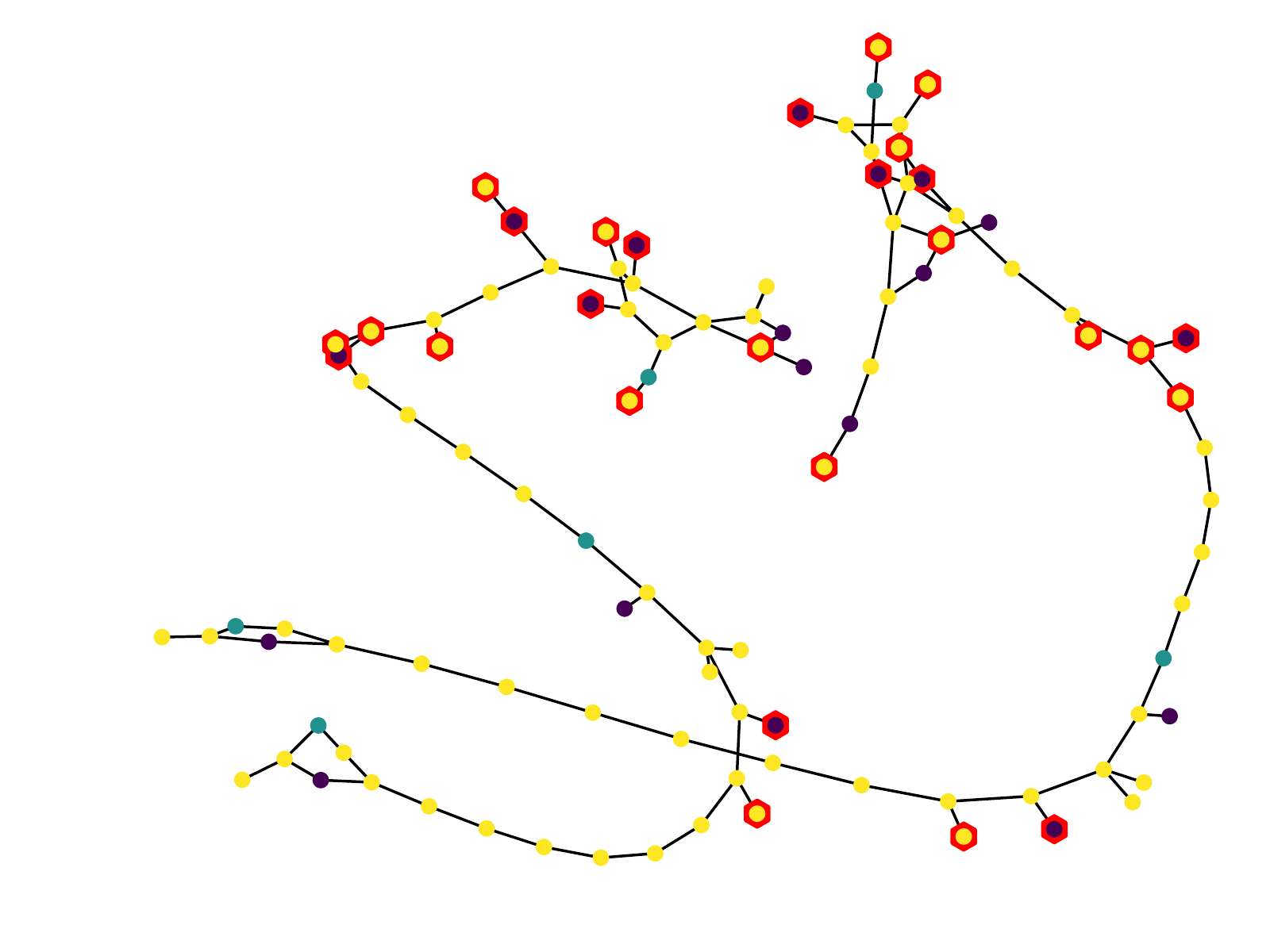}
\end{minipage}
}
\\
\subfigure[Local pooling ($k=2$).]{
\begin{minipage}{0.48\linewidth}
\centering
\includegraphics[width=2.5in]{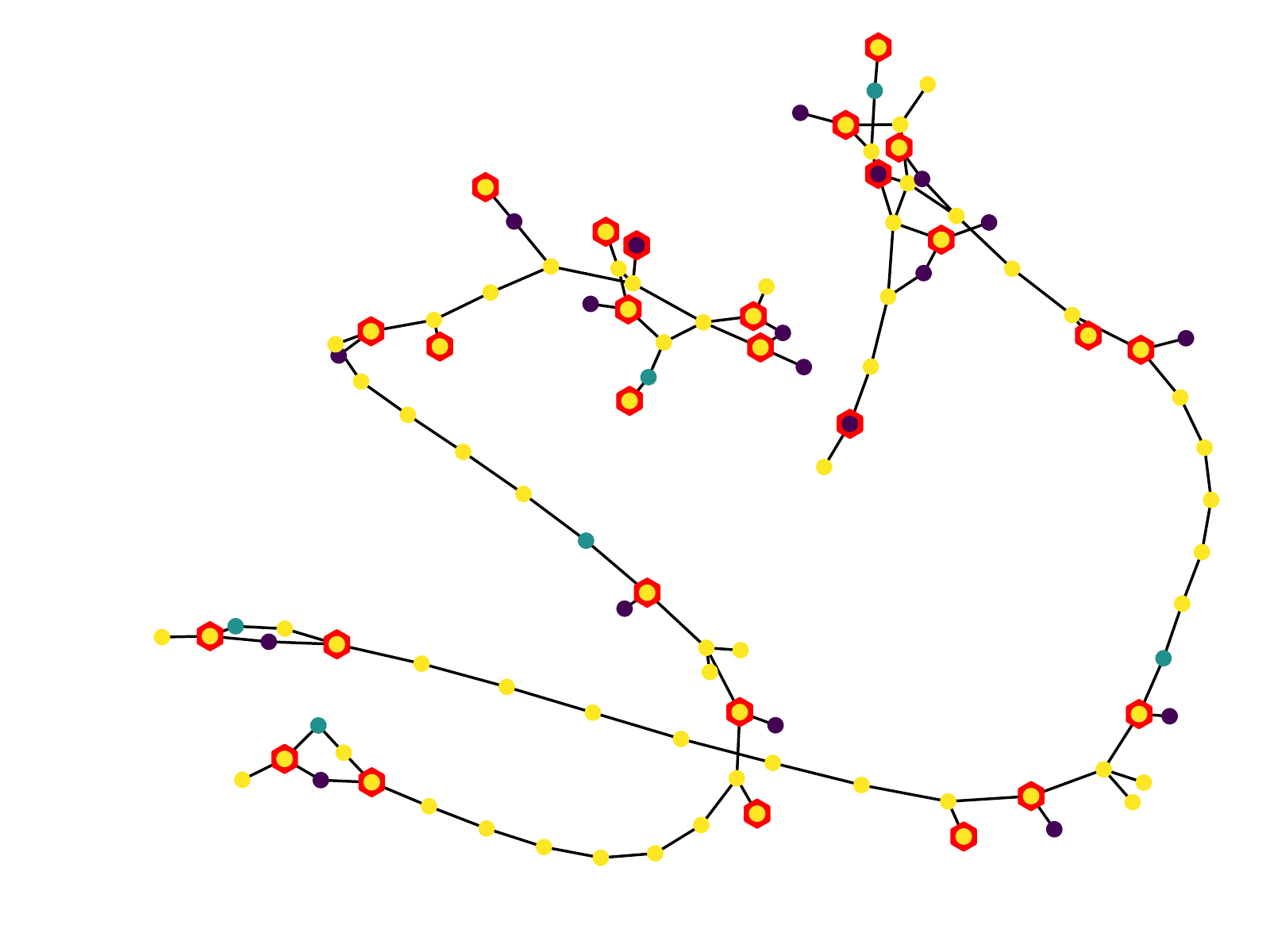}
\end{minipage}
}
\subfigure[ DIFFPOOL. ]{
\begin{minipage}{0.48\linewidth}
\centering
\includegraphics[width=2.5in]{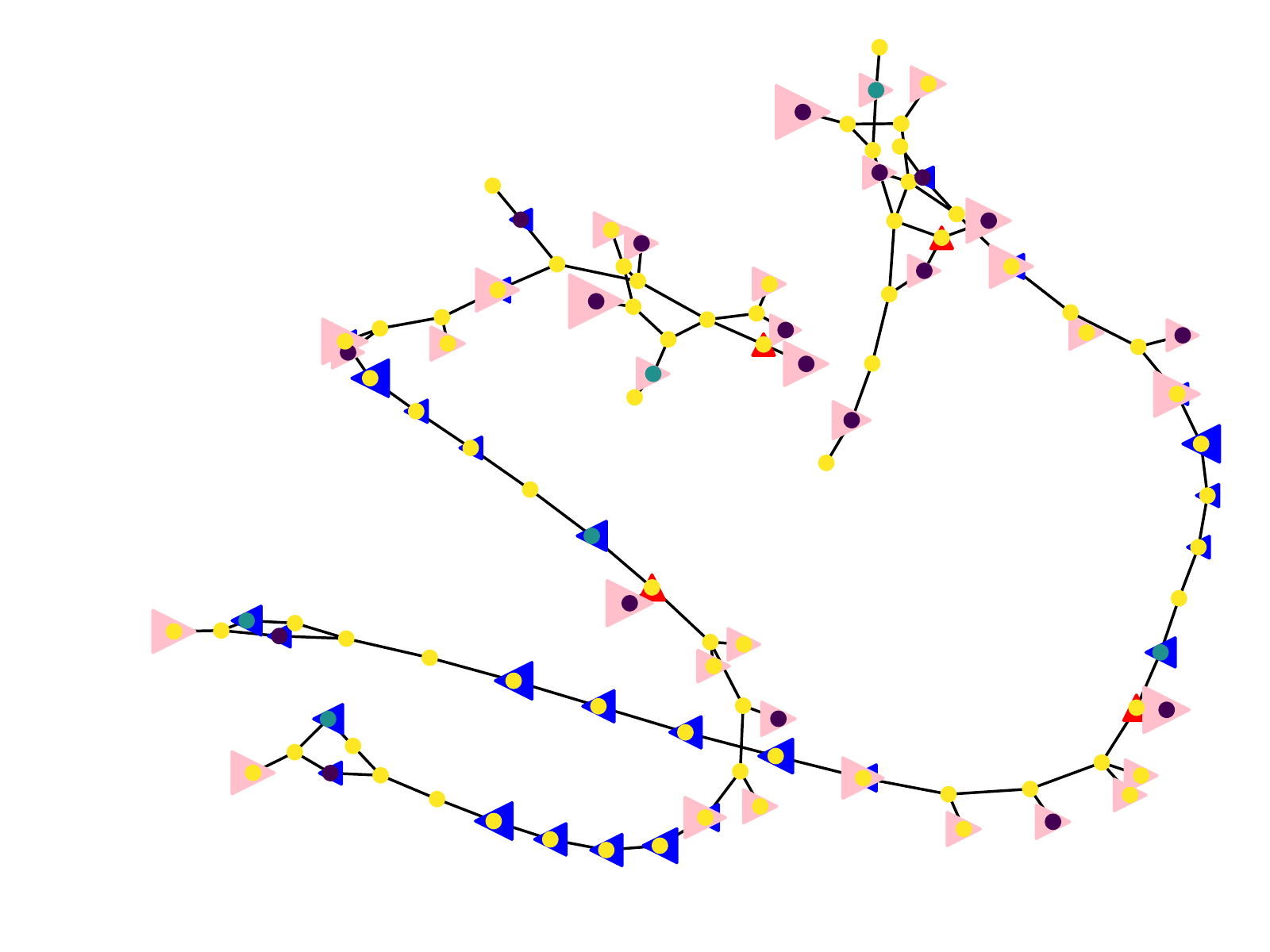}
\end{minipage}
}
\centering
\caption{ Illustration of different pooling schemes with  the color of the node indicating the node category on a graph of the NCI1 dataset.  Results of global and local strategies of iPool are respectively shown in (a)$\sim$(c), where nodes with red hexagons are  ones transferred to the coarsened graph. The result of the DIFFPOOL is shown in (d), where  only first three nodes in the coarsened graph are presented and only the nodes with assign ratio larger than 0.1  be absorbed to the new node are illustrated for clarity.  The triangular nodes  with the same color  are softly assigned to the same node in the coarsened graph with the size of the triangular being proportional to the assign ratio.}\label{fig:3}
\end{figure}

\begin{table}[tp]
\centering
\caption{ Running time (s/epoch).} \label{t:3}
 \begin{tabular}{l|cccccc}
 \toprule
	Method & NCI1 &NCI109&MUTAG&IMDB-B&IMDB-M& 
Multiple\\
	\midrule
 	SET2SET& 24.86 (7.1)&26.08 (7.1)&0.45 (2.0)&7.71 (7.1)&8.25 (7.1)&$\times$7.1\\
 	DIFFPOOL&4.03 (1.1)&4.05 (1.1)&0.25 (1.1)&1.25 (1.1)&1.30 (1.1)&$\times$1.1\\
 	Proposed&3.51&3.66&0.22&1.09&1.16&1\\
 	\bottomrule
 \end{tabular}
 \end{table}

\subsection{Experimental results and analysis}

\begin{figure}[tp]
\centering
\subfigure[D\&D]{
\begin{minipage}{0.48\linewidth}
\centering
\includegraphics[width=2.8in]{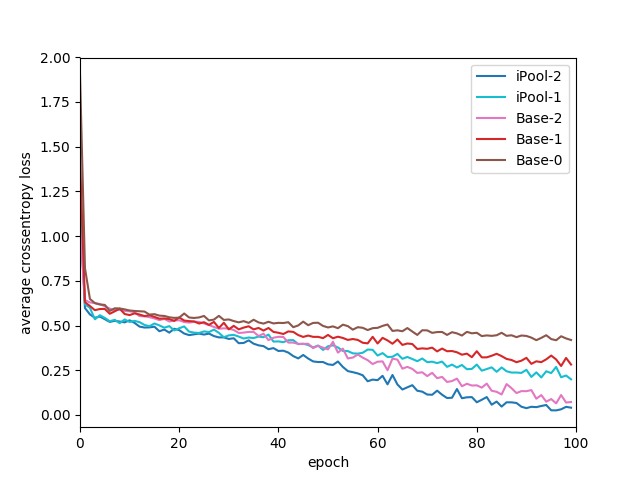}
\end{minipage}
}
\subfigure[Reddit-Multi-12k]{
\begin{minipage}{0.48\linewidth}
\centering
\includegraphics[width=2.8in]{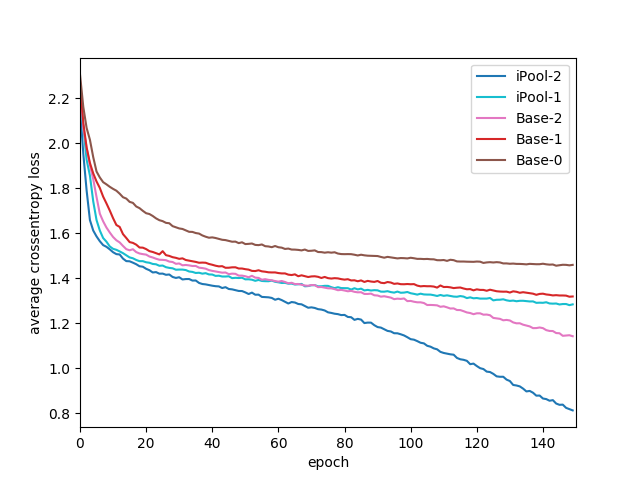}
\end{minipage}
}
\centering
\caption{ Training convergence on the D\&D and REDDIT-MULTI-12K datasets.}\label{fig:5}
\end{figure}
\begin{table}[tp]
\centering
\caption{ Results of ablation experiments.} \label{t:4}
\begin{tabular}{l|ccccc}
 \toprule
Dataset& Base-0 & Base-1 & Base-2 & iPool-1& iPool-2  \\
\midrule
D\&D& 78.52$\pm$3.89&76.37$\pm$3.13&75.94$\pm$4.47&79.28$\pm$2.29&79.45$\pm$2.78 \\
 RED-M-12K&46.85$\pm$1.13&48.44$\pm$1.23&48.58$\pm$1.65&48.71$\pm$0.95&47.64$\pm$1.56\\
\bottomrule
\end{tabular}
\end{table}

Table~\ref{t:1} and Table~\ref{t:2} respectively illustrate performance of two kinds of networks, a small one (30 hidden neurons) and a large one (64 hidden neurons). The proposed iPool strategies outperform other GNNs based methods on 9 of 10 datasets. Compared with the state-of-the-art  kernel based methods, the iPool methods also achieve competitive performance. The limited number of training samples per category would cause the neural network overfitting, which probably explains the inferior performance of GNNs based methods on the ENZYMES dataset. With regards to the COLLAB dataset, the WL kernel and WL-OA kernel outperforms all GNN variants, which probably results from the large number of node degrees  making it difficult for graph neural networks to effectively aggregate neighborhood information with convolution layers. 

We also note that the local pooling strategy performs better than the global one on all of the datasets. The local strategy prefers nodes from different neighborhoods and thereby the coarsened graphs could better preserve the structure of the original graphs, as demonstrated in Fig.~\ref{fig:3}.  In addition, the number of hops (i.e., $k$) utilized in the prediction function (Eq. (\ref{e:2})) has an important impact on the classification performance, and has a close relationship with the average node degree. Specifically, the sparser the connections between nodes of a graph, the larger the necessary number of hops ($k$) to achieve the best performance. This is consistent with intuition, since the prediction function with large $k$ in graphs with dense connections, will aggregate global information of the whole graph rather than local information of each neighborhood. 
 
Without the graph pooling layer, the GraphSAGE and SET2SET methods extract representations of graphs at the scale of the original signal. In contrast,  DIFFPOOL and iPool obtain multiscale graph representations and achieve better results on most datasets. In addition, the iPool scheme further outperforms DIFFPOOL on all of the datasets. It probably better preserves the structure of the original graphs and the localization property of graph signals, as demonstrated in Fig.~\ref{fig:3}.  In addition, the running speed of iPool is about 7 times faster than SET2SET and 1.1 times faster than DIFFPOOL, as illustrated in Table~\ref{t:3}.
 
\textbf{Ablation studies.} We further explore the effectiveness of the iPool operator to deal with multiscale representations of graph data with a series of ablation experiments on the D\&D and REDDIT-MULTI-12K datasets,  the largest datasets of bioinformatics and social networks respectively in terms of  average number of nodes per graph. Specifically, we conduct experiments on 3 network architectures, as presented in  Fig.~\ref{fig:1}, to deal with single-scale (Base-0, 1 convolution module), double-scale (iPool-1, 2 convolution modules and 1 pooling module), and triple-scale  (iPool-2, 3 convolution modules and 2 pooling modules) graph representations.  To eliminate the impact of the depth of neural networks, we  further extract single-scale graph representations with the same architecture as their multiscale counterparts  except for the pooling layers, i.e., Base-1 corresponding to iPool-1 and  Base-2 matching iPool-2. We utilize the same procedure as the one for Table.~\ref{t:1} and report 10-fold cross-validation results in terms of the mean and standard  variation of classification accuracy in Table~\ref{t:4}.

According to Fig.~\ref{fig:5} and Table~\ref{t:4}, we have the following findings. (1) The iPool operation  could accelerate the convergence of representation learning in the training phase. Concretely, the iPool-2 and iPool-1 models converge faster than their single-scale counterparts Base-2 and Base-1, and are far faster than the single-scale  shallow model Base-0 on both datasets. (2)  The iPool operation could improve the classification performance of models due to its hierarchical representation. Specifically, models with iPool achieve the best performance on both datasets. Compared to iPool-1, the inferior performance of iPool-2 on the REDDIT-MULTI-12K dataset  probably results from the overfitting of the model, given  the simple constant node signals and the scaled-down graphs through two pooling operators.

\section{Conclusion}
We have proposed in this paper a low complexity and adaptive graph pooling operator for GNNs, which improves their capability of distilling hierarchical representations of graphs and network data. The new operator has interesting properties in practice, in that it is mostly based on local computations, and it leads to invariance properties under graph isomorphism. The proposed iPool solution further permits to achieve state-of-the-art performance on several graph classification datasets. An interesting future direction is to explore other neighborhood prediction functions to make the neighbor information gain in line with more general conditional distribution of nodes. It is also worthwhile to utilize the proposed pooling operation with other convolution schemes, such as GINs. 

\bibliographystyle{abbrvnat}
\bibliography{egbib}

 \section*{Appendix}
 \renewcommand{\thesection}{A\arabic{section}}
 \renewcommand{\thetable}{A\arabic{table}}
  \renewcommand{\thefigure}{A\arabic{figure}}
    \renewcommand{\theequation}{A\arabic{equation}}
  \setcounter{section}{0}
 \setcounter{table}{0}
  \setcounter{figure}{0}
    \setcounter{equation}{0}
 \section{Proof of Proposition 1}
\begin{proof}
\begin{subequations}
\renewcommand{\theequation}{\theparentequation.\arabic{equation}}
\begin{align}
 &H(\mathbf{x}_{l,i}| \{\mathbf{x}_{l,j}\}_{  N(v_{l,i})}) \nonumber \\
 	=&\mathbb{E}[-\log p(\mathbf{x}_{l,i}| \{\mathbf{x}_{l,j}\}_{  N(v_{l,i})})] \nonumber \\
 	=&\mathbb{E}[-\log \Pi _{z=1}^{d} p(x_{l,i,z}| \{\mathbf{x}_{l,j}\}_{  N(v_{l,i})})]  \label{e:a1.1} \\
\approx & \frac{1}{m}\sum_{k=1}^{m} -\log  \Pi _{z=1}^{d} p(x_{l,i,z}^{(k)}| \{\mathbf{x}_{l,j}^{(k)}\}_{ N(v_{l,i})})  \label{e:a1.2} \\
 =&\frac{1}{mb_{l,i} }\sum_{k=1}^{m} \sum_{z=1}^{d}|x_{l,i,z}^{(k)}-\mu_{l,i,z}^{(k)}|+d\log2b_{l,i} \nonumber \\
  =&\frac{1}{mb_{l,i} }\sum_{k=1}^{m} \parallel \mathbf{x}_{l,i}^{(k)}-\mathbf{\mu}_{l,i}^{(k)}\parallel_{1}+d\log2b_{l,i} \nonumber\\
 \approx & \frac{1}{b_{l,i} } \parallel \mathbf{x}_{l,i}-\mathbf{\mu}_{l,i}\parallel_{1}+d\log2b_{l,i} \label{e:a1.3} \\
  =&\frac{1}{b_{l,i} } \parallel \mathbf{x}_{l,i}-f(v_{l,i})\parallel_{1}+ d\log2b_{l,i}\nonumber  \\
    =&\frac{1}{b_{l,i} }\gamma(v_{l,i})+d\log2b_{l,i},
\end{align}
\end{subequations}
where Eq.~(\ref{e:a1.2}) utilizes an empirical probability to approximate Eq.~(\ref{e:a1.1}) and Eq.~(\ref{e:a1.3})  further approximates it with one sample estimation.
\end{proof}
\section{Proof of Proposition 2}

\begin{proof}
Since $\mathcal{G}_{l} \simeq  \mathcal{G}_{l}^{'}$, there exists an edge-preserving bijection:
\begin{equation}
	t: \mathcal{V}_{l} \longrightarrow \mathcal{V}_{l}^{'}.
\end{equation}
For $\forall  v_{l,i} \in \mathcal{V}_{l}$,  there is a $v_{l,m}^{'}=t(v_{l,i}) \in \mathcal{V}_{l}^{'}$ and their neighborhood information gains are respectively:
\begin{gather}
	\gamma(v_{l,i})	=\parallel x(v_{l,i})-\frac{1}{k}\sum_{h=1}^{k}\sum_{v_{l,j} \in N^{h}(v_{l,i})}(\bar{{P_{l}}^{h}})_{ij} \times x(v_{l,j}) \parallel_{1}.\\
\gamma(v_{l,m}^{'})	=\parallel x(v_{l,m}^{'})-\frac{1}{k}\sum_{h=1}^{k}\sum_{v_{l,n}^{'} \in N^{h}(v_{l,m}^{'})}(\bar{{P_{l}^{'}}^{h}})_{mn}\times x(v_{l,n}^{'}) \parallel_{1}.
\end{gather}
Since $t(\cdot)$ is edge-preserving, $N^{h}(v_{l,i})=N^{h}(v_{l,m}^{'})$ and any edge $(v_{l,i}, v_{l,j}) \in  \mathcal{E}_{l}$ shares the same weights with its counterpart $(v_{l,m}^{'}, v_{l,n}^{'}) \in  {\mathcal{E}}_l^{'}$:
\begin{equation}\label{e:p1}
	(A_{l})_{ij}=(A_l^{'})_{mn},\quad ({A_l}^{h})_{ij}=({A_l^{'}}^{h})_{mn}, \quad (\bar{{P_{l}}^{h}})_{ij}=(\bar{{P_{l}^{'}}^{h}})_{mn}.
\end{equation}
Therefore, 
\begin{gather}
	\gamma(v_{l,i})=\gamma(v_{l,m}^{'}), \quad  \forall  v_{l,i} \in \mathcal{V}_{l},  
\end{gather}
and it holds true also for the normalized neighborhood information gain. Note that the ranking function takes only the value of (normalized) neighborhood information gain of nodes under consideration, and that $v_{l,i}$ has the same ranking as $t(v_{l,i})$. Then
\begin{equation}
	 \mathcal{V}_{l+1}= \mathcal{V}_{l+1}^{'},  \quad  \mathcal{E}_{l+1}= \mathcal{E}_{l+1}^{'}, \quad  X_{l+1}=X_{l+1}^{'}.
\end{equation}

Thus, the iPool operation is invariant to isomorphism.
\end{proof}

\section{More information about experiments}
\begin{table}[htp]
    \centering
    \caption{Hyper-paramters used in the experiments. The `lr'  denotes learning rate and the `k-global' indicates the number of hops used in the prediction function under the global iPool strategy. } \label{t:a1}
    \begin{tabular}{c|ccccc}
    \toprule
         Dataset  &lr   &s     &k-global  & drop-out   &weight decay   \\
         \hline
         ENZY     &1e-3 &2     &1         &  0       &1e-4 \\
         D\&D     &1e-4 &2     &1         &0.5       &0\\
         RED-M-12K&1e-3    &2     &2         &0         &0\\
         COLL     &1e-3 &1     &1         &0         &0\\
         PROT     &1e-2 &2     &1         &0         &1e-4\\
         NCI1     &1e-3 &2     &2         &0         &1e-4 \\
         NCI109   &1e-3 &2     &2         &0.5       &0\\
         MUTAG    &1e-2 &2     &2         &0.5    &3e-5\\
         IMDB-B   &1e-3 &1     &2         &0         &1e-4\\
         IMDB-M   &1e-3 &1     &2         &0         &0\\
         \bottomrule
    \end{tabular}
    \label{tab:my_label}
\end{table}

\end{document}